\pdfoutput=1

\documentclass[11pt]{article}

\usepackage[]{acl}
\usepackage[normalem]{ulem}
\usepackage{times}
\usepackage{latexsym}
\usepackage{algorithm}
\usepackage{adjustbox}
\usepackage{mathtools}
\usepackage{graphicx}
\usepackage{float}
\usepackage{subfigure}
\usepackage{caption}
\usepackage{algpseudocode}
\usepackage{soul}
\usepackage[T1]{fontenc}

\usepackage[utf8]{inputenc}
\usepackage{multirow}
\usepackage{microtype}
\usepackage{fixltx2e}
\newcommand\crule[3][black]{\textcolor{#1}{\rule{#2}{#3}}}
\usepackage[prependcaption,textsize=tiny]{todonotes}
\definecolor{ao}{rgb}{0.0, 0.5, 0.0}
\definecolor{dark}{rgb}{0.20784313725,0.10980392156,0.45882352941}
\definecolor{med}{rgb}{0.40392156862,0.30588235294,0.65490196078}
\definecolor{light}{rgb}{0.70588235294,0.65490196078,0.83921568627}
\definecolor{light_green}{rgb}{0.85098039215,0.91764705882,0.82745098039}
\definecolor{special_red}{rgb}{0.59607843137,0.59607843137,0.59607843137}
\definecolor{g1}{rgb}{0.36078431372,0.65098039215,0.2862745098}
\definecolor{g2}{rgb}{0.26666666666,0.50588235294,0.55294117647}

\title{TransLIST: A Transformer-Based Linguistically Informed Sanskrit Tokenizer}

\newcommand*{\affmark}[1][*]{\textsuperscript{#1}}

\author{Jivnesh Sandhan\affmark[1], Rathin Singha\affmark[2], Narein Rao\affmark[1], Suvendu Samanta\affmark[1], \\\textbf{Laxmidhar Behera\affmark[1,4] and Pawan Goyal\affmark[3]}\\
\affmark[1]IIT Kanpur, \affmark[2]UCLA, \affmark[3]IIT Kharagpur, \affmark[4]IIT Mandi\\
\texttt{jivnesh@iitk.ac.in,rsingha108@g.ucla.edu,}\\
\texttt{nrao20@iitk.ac.in,pawang@cse.iitkgp.ac.in}}

\begin{document}
\maketitle
\begin{abstract}
Sanskrit Word Segmentation (SWS) is essential in making digitized texts available and in deploying downstream tasks.
It is, however, non-trivial because of the {\sl sandhi} phenomenon that modifies the characters at the word boundaries, and needs special treatment. Existing \textit{lexicon driven} approaches for SWS make use of Sanskrit Heritage Reader, a lexicon-driven shallow parser, to generate the complete candidate solution space, over which  various methods are applied to produce the most valid solution. However, these approaches fail while encountering out-of-vocabulary tokens. On the other hand, \textit{purely engineering} methods for SWS have made use of recent advances in deep learning, but cannot make use of the latent word information on availability. 

To mitigate the shortcomings of both families of approaches, we propose \textbf{Tran}sformer based \textbf{L}inguistically \textbf{I}nformed \textbf{S}anskrit \textbf{T}okenizer (TransLIST) consisting of (1) a module that encodes the character input along with latent-word information, which takes into account the \textit{sandhi} phenomenon specific to SWS and is apt to work with partial or no candidate solutions, (2) a novel soft-masked attention to prioritize potential candidate words and (3) a novel path ranking algorithm to rectify the corrupted predictions.  Experiments on the benchmark datasets for SWS show that TransLIST outperforms the current state-of-the-art system by an average 7.2 points absolute gain in terms of perfect match (PM) metric.\footnote{The codebase and datasets are publicly available at: \url{https://github.com/rsingha108/TransLIST}} 

\end{abstract}

\section{Introduction}
\begin{figure}[!thb]
\centering
\includegraphics[width=0.4\textwidth]{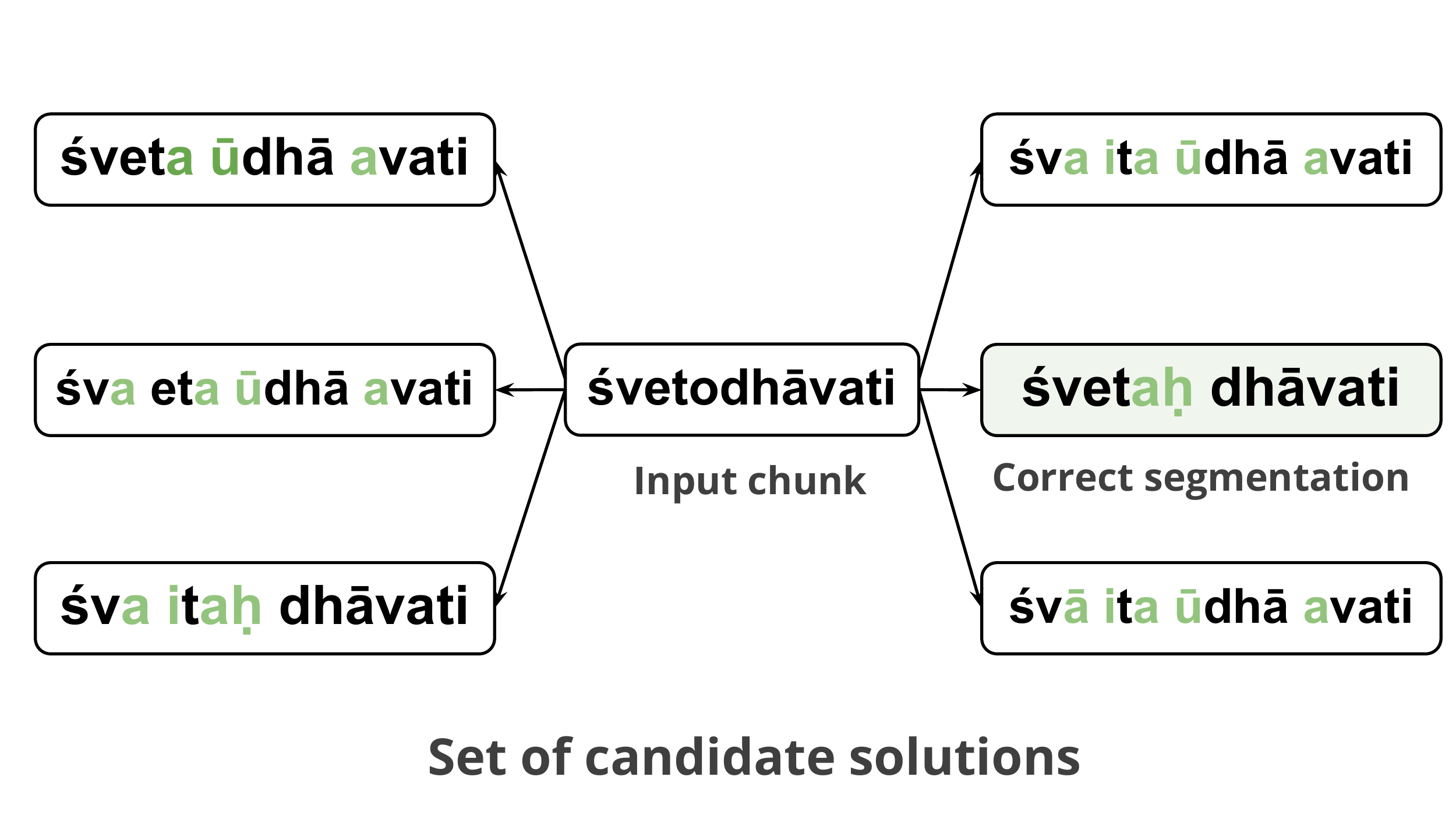}
\caption{An example to illustrate challenges posed by \textit{sandhi} phenomenon for SWS task.} 
\label{fig:sandhi_issue} 
\end{figure}

Sanskrit is considered as a cultural heritage and knowledge preserving language of ancient India. The momentous development in digitization efforts has made ancient manuscripts in Sanskrit readily available for the public domain. However, the usability of these digitized manuscripts is limited due to linguistic challenges posed by the language. SWS conventionally serves the most fundamental prerequisite for text processing step to make these digitized manuscripts accessible  and to deploy many downstream tasks such as text classification \cite{sandhan-etal-2019-revisiting,krishna-etal-2016-compound}, morphological tagging \cite{gupta-etal-2020-evaluating,krishna-etal-2018-free}, dependency parsing \cite{sandhan-etal-2021-little,krishna-etal-2020-keep}, automatic speech recognition \cite{kumar2022linguistically} etc. SWS is not straightforward due to the phenomenon of \textit{sandhi}, which creates phonetic transformations at word boundaries. This not only obscures the word boundaries but also modifies the characters at juncture point by deletion, insertion and substitution operation. Figure~\ref{fig:sandhi_issue} illustrates some of the syntactically possible splits due to the language-specific \textit{sandhi} phenomenon for Sanskrit. This demonstrates the challenges involved in identifying the location of the split and the kind of transformation performed at word boundaries. 

The recent surge in SWS datasets \cite{krishna-etal-2017-dataset,hackathon_data} has led to various methodologies to handle SWS. Existing {\sl lexicon-driven} approaches rely on a \textit{lexicon driven} shallow parser, popularly known as Sanskrit Heritage Reader (SHR) \cite{goyal2016}.\footnote{https://sanskrit.inria.fr/DICO/reader.fr.html} This line of approaches~\cite{krishna-etal-2016-word,krishna-etal-2018-free,amrith21} formulate the task as finding the most accurate semantically and syntactically valid solution from the candidate solutions generated by SHR.
With the help of the significantly reduced exponential search space provided by SHR and linguistically involved feature engineering, these \textit{lexicon driven} systems
\cite{amrith21,krishna-etal-2018-free} report close to state-of-the-art performance for the SWS task. However, these approaches rely on the completeness assumption of SHR, which is optimistic given that SHR does not use domain specific lexicons. These models are handicapped by the failure of this preliminary step. On the other hand, \textit{purely engineering} based knowledge-lean data-centric approaches \cite{hellwig-nehrdich-2018-sanskrit,reddy-etal-2018-building,aralikatte-etal-2018-sanskrit} perform surprisingly well without any explicit hand-crafted features and external linguistic resources. These \textit{purely engineering} based approaches are known for their ease of scalability and deployment for training/inference. However, a drawback of these approaches is that they are blind to latent word information available through external resources.

There are also lattice-structured approaches \cite{zhang-yang-2018-chinese,gui-etal-2019-lexicon,li-etal-2020-flat} (originally proposed for Chinese Named Entity Recognition (NER), which incorporate lexical information in character-level sequence labelling architecture). However, these approaches cannot be directly applied for SWS; since acquiring word-level information is not trivial due to \textit{sandhi} phenomenon. To overcome these shortcomings, 
  we propose \textbf{Trans}former-based \textbf{L}inguistically \textbf{I}nformed \textbf{T}okenizer (TransLIST).
TransLIST is a perfect blend of \textit{purely engineering} and \textit{lexicon driven} approaches for the SWS task and provides the following advantages: (1) Similar to \textit{purely engineering} approaches, it facilitates ease of scalability and deployment during training/inference. (2) Similar to \textit{lexicon driven} approaches, it is capable of utilizing the candidate solutions generated by SHR, which further improves the performance. (3) Contrary to \textit{lexicon driven} approaches, TransLIST is robust  and can function even when candidate solution space is partly available or unavailable. 

Our key contributions are as follows:
(a) We propose the linguistically informed tokenization module (\S~\ref{LIST}) which accommodates language-specific \textit{sandhi} phenomenon and adds inductive bias for the SWS task.
(b) We propose a novel soft-masked attention (\S~\ref{sma_subsection}) that helps to add inductive bias for prioritizing potential candidates keeping mutual interactions between all candidates intact.
(c) We propose a novel path ranking algorithm (\S~\ref{constrained_inference}) to rectify the corrupted predictions. 
(d) We report an average 7.2 points perfect match absolute gain (\S~\ref{results}) over the current state-of-the-art system \cite{hellwig-nehrdich-2018-sanskrit}.

We elucidate our findings by first describing TransLIST and its key components (\S~\ref{system_architecture}), followed by the evaluation of TransLIST against strong baselines on a test-bed of 2 benchmark datasets for the SWS task (\S~\ref{experiments}). Finally, we investigate and delve deeper into the capabilities of the proposed components and its corresponding modules (\S~\ref{fine_grained_analysis}).

\begin{figure*}[!tbh]
    \centering
    \subfigure[]{\includegraphics[width=0.45\textwidth]{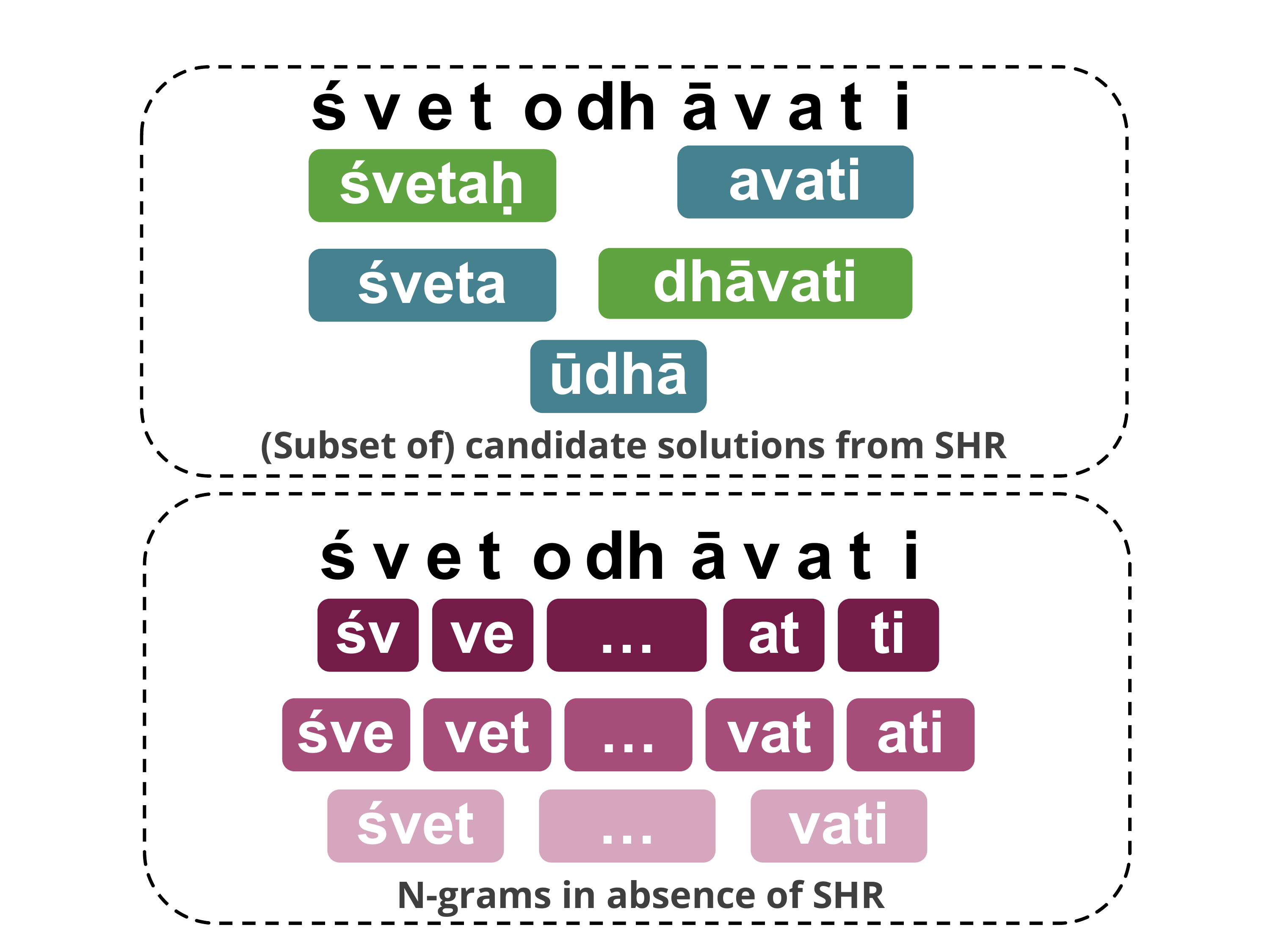}\label{fig:SHR2lattice}} 
    \subfigure[]{\includegraphics[width=0.45\textwidth]{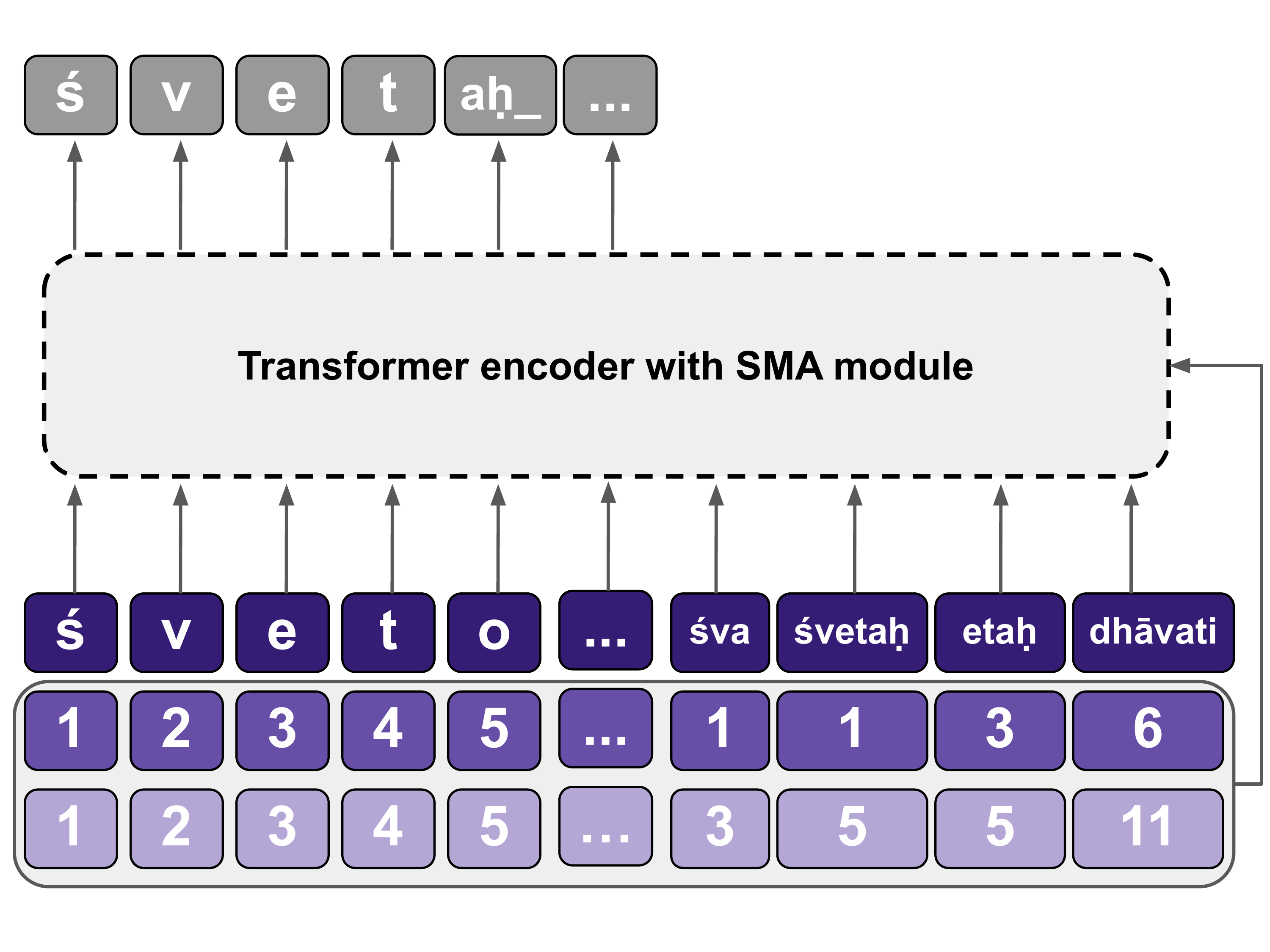}\label{fig:flat_architecture}} 
    \caption{Illustration of TransLIST with a toy example ``{\sl \'{s}vetodh\={a}vati}''. Translation: ``The white (horse) runs.'' (a) LIST module: We use the candidate solutions (two possible candidate solutions are highlighted with \crule[g1]{0.25cm}{0.25cm},\crule[g2]{0.25cm}{0.25cm} colors where the latter is the gold standard) from SHR if available; in the absence of SHR, we resort to using n-grams ($n \leq 4$). (b) TransLIST architecture: In span encoding, each node is represented by head and tail position index of its character in the input sequence. \crule[dark]{0.25cm}{0.25cm}, \crule[med]{0.25cm}{0.25cm}, \crule[light]{0.25cm}{0.25cm} denote tokens, heads and tails, respectively. The SHR helps to include words such as \textit{\'{s}va, \'{s}veta\d{h}} and \textit{eta\d{h}} whose boundaries are modified with respect to input sequence due to \textit{sandhi} phenomenon. Finally, on the top of the Transformer encoder, classification head learns to predict gold standard output shown by \crule[special_red]{0.25cm}{0.25cm} for the corresponding input character nodes only.}
    \label{figure:complete_TransLIST}
\end{figure*}

\section{Methodology}
In this section, we will examine the key components of TransLIST which includes a linguistically informed tokenization module that encodes character input with latent-word information while accounting for SWS-specific \textit{sandhi} phenomena (\S~\ref{LIST}), a novel soft-masked attention to prioritise potential candidate words (\S~\ref{sma_subsection}) and a novel path ranking algorithm to correct mispredictions (\S~\ref{constrained_inference}).
\label{system_architecture}
\subsection{Linguistically Informed Sanskrit Tokenizer (LIST)}
\label{LIST}
\textit{Lexicon driven} approaches for SWS are brittle in realistic scenarios and \textit{purely engineering} based approaches do not consider the potentially useful latent word information. We propose a win-win/robust solution by formulating SWS as a character-level sequence labelling integrated with latent word information from the SHR as and when available.
TransLIST is illustrated with an example \textit{\`{s}vetodh\={a}vati} in Figure~\ref{figure:complete_TransLIST}. SHR employs a Finite State Transducer (FST) in the form of a lexical juncture system to obtain a compact representation of candidate solution space aligned with the input sequence. As shown in Figure~\ref{fig:SHR2lattice}, we receive the candidate solution space from the SHR engine. Here, \textit{\`{s}vetah dh\={a}vati} and \textit{\`{s}veta \={u}dh\={a} avati} are two syntactically possible splits.\footnote{Only some of the solutions are shown for visualization.} It does not suggest the final segmentation. The candidate space includes words such as \textit{\`{s}va, \`{s}veta\d{h}} and \textit{eta\d{h}} whose boundaries are modified with respect to the input sequence due to \textit{sandhi} phenomenon. SHR gives us mapping (head and tail position) of all the candidate nodes with the input sequence. 
In case such mapping is incorrect for some cases, we rectify it with the help of deterministic algorithm by matching candidate nodes with the input sentence and finding the closest match.
In the absence of SHR, we propose to use all possible n-grams ($n \leq 4$)\footnote{We do not observe significant improvements for $n > 4$.}which helps to add inductive bias about neighboring candidates in the window size of 4.\footnote{Our probing analysis (Figure~\ref{fig:sma_probing}) suggests that char-char attention mostly focuses on immediate neighbors. Refer to \S~\ref{fine_grained_analysis} for detailed ablations on LIST variants.}
We feed the candidate words/n-grams to the Transformer encoder and the classification head learns to predict gold standard output for the corresponding input character nodes only.
The output vocabulary consists of unigram characters (e.g., \'{s}, v), bigrams and tri-grams (e.g., a\d{h}\_). The output vocabulary contains `\_' to represent spacing between words.
Consequently, TransLIST is capable of using both character-level modelling as well as latent word information as and when available. On the other hand, \textit{purely engineering} approaches rely only on character-level modelling  and \textit{Lexicon driven} approaches rely only on word-level information from SHR to handle \textit{sandhi}.
\subsection{Soft Masked Attention (SMA)}
\label{sma_subsection}
Transformers \cite{vasvani_attention} have been proven to be effective for capturing long-distance dependencies in a sequence. The self-attention property of a Transformer facilitates effective interaction between character and available latent word information. There are two preliminary prerequisites for effective modelling of inductive bias for tokenization: (1) Allow interactions between the candidate words/characters within and amongst chunks. (2) Prioritize candidate words containing the input character for which a prediction is being made (e.g., in Figure~\ref{fig:flat_architecture}, \textit{\`{s}va} and \textit{\`{s}veta\d{h}} are prioritized amongst the candidate words when predicting for the character \textit{\`{s}}).\footnote{We find that failing to meet any one of the prerequisites leads to drop in performance (\S~\ref{fine_grained_analysis}).} 
The vanilla self-attention \cite{vasvani_attention} can address both the requirements; however, it has to self-learn the inductive bias associated with prioritisation. It may not be an effective solution in low-resourced settings. On the other hand, if we use hard-masked attention to address the second prerequisite, we lose mutual interactions between the candidates. Hence, we propose a novel soft-masked attention which helps to address both the requirements effectively. To the best of our knowledge, there is no existing soft-masked attention similar to ours. We formally discuss this below.

\textbf{Self-attention} maps a query and a set of key-value pairs to an output as discussed in \newcite{vasvani_attention}. For an input $x=(x_1, ..., x_n)$ where $x_i\in R^{d_x}$, self-attention gives an output $z=(z_1, ..., z_n)$ where $z_i\in R^{d_z}$. We presume the standard formulation of vanilla self-attention \cite{vasvani_attention} where $d_x$ is the dimension of input word representation and $d_z$ is the projection dimension. Here, $W^Q,W^K,W^V \in R^{d_x \times d_z}$ are parameter matrices. For simplicity, we ignore multi-head attention in equations \ref{eq1}, \ref{eq2} and \ref{eq3}.
\begin{align}
       z_i&= \sum_{j=1}^{n}\alpha_{ij}(x_jW^V)\label{eq1}\\
       \alpha_{ij}&= \frac{\exp{(e_{ij}})}{\sum_{k=1}^{n}\exp{(e_{ik})}}\label{eq2}\\
       e_{ij}&=\frac{(x_iW^Q)(x_jW^K)^T}{\sqrt{d_z}}\label{eq3}
\end{align}

In \textbf{soft-masked attention}, we provide a prior about interactions between candidate words and the  input characters using a span encoding ($s_{ij} \in R^{d_z}$) \cite{li-etal-2020-flat}. Intuitively, it helps inject inductive bias associated with prioritisation whilst maintaining mutual interactions between the candidates.

Formally, we modify Equation \ref{eq2} to define soft masked attention as:
\begin{align}
       \alpha^{SM}_{ij}&=\frac{M_{ij}\exp{(e_{ij}})}{\sum_{k=1}^{n}M_{ik}\exp{(e_{ik})}}
\end{align}
where $M \in R^{n \times n}$, $M_{ij}\in [0,1]$. 
$M_{ij}$ is defined as:
\begin{align}
       M_{ij}&=\frac{(x_iW^Q)(s_{ij}W^R)^T}{\sqrt{d_z}}
\end{align}
 $W^R \in R^{d_z \times d_z}$ is a learnable parameter which projects $s_{ij}$ into a location-based key vector space. Summarily, the proposed SMA module helps to prioritize potential candidate words with the help of separation, inclusion and intersection information between nodes. Finally, we calculate the output $z$ with the help of the proposed SMA as follows:
\begin{align}
       z_i&= \sum_{j=1}^{n}\alpha^{SM}_{ij}(x_jW^V)
\end{align}
Next, we discuss the span position encoding.

\textbf{Span position encoding} is one of the backbones of the proposed soft-masked module. It is utilized to capture the interactions between the candidate words and the sequence of input characters. Each span/node (which is a character/word and its corresponding position in the input sentence) is represented by the head and tail which denote the position index of the initial and final characters of the token in the input sequence, as shown in Figure~\ref{fig:flat_architecture}. The span of character is characterized by the same head and tail position index. For example, $head[i]$ and $tail[i]$ represent the head and tail index of span $x_i$, respectively. The separation, inclusion and intersection information between nodes $x_i$ and $x_j$ can be captured by  the four distance equations \ref{eq7}-\ref{eq10}.

\begin{align}    
       d_{ij}^{(hh)}&=head[i]-head[j]\label{eq7}\\
       d_{ij}^{(ht)}&=head[i]-tail[j]\label{eq8}\\
       d_{ij}^{(th)}&=tail[i]-head[j]\label{eq9}\\
       d_{ij}^{(tt)}&=tail[i]-tail[j]\label{eq10}
\end{align}
 The final span encoding is a non-linear transformation of these 4 distances:
\begin{align}
        s_{ij}&=\text{ReLU}(w_s(p_{d_{ij}^{(hh)}}\oplus p_{d_{ij}^{(ht)}}\oplus p_{d_{ij}^{(th)}}\oplus p_{d_{ij}^{(tt)}}))
\end{align}
where $w_s \in R$ is a learnable parameter, $\oplus$ is a concatenation operation and $p_d \in R^{\frac{d_z}{4}}$ is a sinusoidal position encoding similar to \newcite{vasvani_attention}.

\subsection{Path Ranking for Corrupted Predictions (PRCP)}
\label{constrained_inference}
Our error analysis (\S~\ref{fine_grained_analysis}) suggests that sometimes the proposed system predicts words that are not part of the candidate solution space. These mistakes can be rectified with the help of SHR's candidate solutions by appropriately substituting suitable candidates.
We refer to the prediction corresponding to a chunk that does not fall in the candidate solution space, as a {\sl corrupted prediction} and define a {\sl path} as the sequence of characters in a candidate solution for a given input. We enumerate all the possible directed paths (In Figure~\ref{fig:SHR2lattice}, two possible candidate solutions are highlighted with \crule[g1]{0.25cm}{0.25cm},\crule[g2]{0.25cm}{0.25cm} colors) corresponding to the input (with a corrupted prediction) and formulate the task as a {\sl path ranking problem}. While designing the path scoring function (S), we consider the following criteria: (1)  Select a path consisting of semantically coherent candidate words. We use an integrated judgment from two sources.  First, we prefer a path having a high log-likelihood (LL) score as per TransLIST to choose a semantically coherent path in line with the contextual information of TransLIST. Second, we reinforce the scoring function (S) by considering the perplexity score ($\rho$) for the path from the character-level language model.
 (2) To avoid paths consisting of over-generated segmentation provided by SHR, we use a penalty proportional to the number of words ($|W|$) present in the path to prefer paths with less number of words.
This gives us the following path scoring function (S):
\begin{equation*}
      S = \frac{LL_{TransLIST}}{\rho_{CharLM} \times |W|}
\end{equation*}
where
\begin{align*}
 LL_{TransLIST} &= \text{log-likelihood by TransLIST}\\
 \rho_{CharLM}    &=  \text{Perplexity score by CharLM} \\ 
 |W| &= \text{Number of words present in path} \\
\end{align*}

\section{Experiments}
\label{experiments}
\paragraph{Data and Metrics:}
\label{data_metrics}
Currently,  Digital Corpus of Sanskrit \cite[DCS]{hellwig2010dcs} has more than 600,000 morphologically tagged text lines. It consists of digitized constructions composed in prose or poetry over a wide span of 3000 years. Summarily, DCS is a perfect representation of various writing styles depending on time and domains. 
We use two available benchmark datasets \cite[SIGHUM]{krishna-etal-2017-dataset}\footnote{\url{https://zenodo.org/record/803508\#.YRdZ43UzaXJ}} and \cite[Hackathon]{hackathon_data} for SWS.
Both datasets are subset of DCS \cite{hellwig2010dcs}.
These datasets also come with candidate solution space generated by SHR for SWS. We prefer \newcite[SIGHUM]{krishna-etal-2017-dataset} over a relatively larger dataset \cite{hellwig-nehrdich-2018-sanskrit} to obviate the time and efforts required for obtaining candidate solution space. We obtain the ground truth segmentation solutions from DCS.  
We could not use DCS10k~\cite{amrith21} due to partly missing gold standard segmentation (inflections) for almost 50\% data points.
SIGHUM consists of 97,000, 3,000 and 4,200 sentences as train, dev, test set, respectively. Similarly, Hackathon consists of 90,000, 10,332 and 9,963 sentences as train, dev and test set, respectively.
We use the following word-level evaluation metrics: macro-averaged Precision (P), Recall (R), F1-score (F) and the percentage of sentences with perfect matching (PM).
 \paragraph{Hyper-parameter settings:} For the implementation of TransLIST, we build on top of codebase by \newcite{li-etal-2020-flat}. We use the following hyper-parameters for the best configuration of TransLIST: number of epochs as
50 and a dropout rate of 0.3 with a learning rate of 0.001. We release our codebase and datasets publicly under the Apache license 2.0. All the artifacts used in this work are publicly available for the research purpose.
For all the systems, we do not use any pretraining. All the input representations are randomly initialized.
 We use GeForce RTX 2080, 11 GB GPU memory computing infrastructure for our experiments.

\paragraph{Baselines:}
\label{baselines}
 We consider two \textit{lexicon-driven} approaches where \newcite[\textbf{SupPCRW}]{krishna-etal-2016-word} formulate SWS as an iterative query expansion problem and   \newcite[\textbf{Cliq-EBM}]{krishna-etal-2018-free} deploy a structured prediction framework.
Next, we evaluate four \textit{purely-engineering} based approaches, namely, Encoder-Decoder framework  \cite[\textbf{Seq2Seq}]{reddy-etal-2018-building},  character-level sequence labelling system with combination of recurrent and convolution element  \cite[\textbf{rcNN-SS}]{hellwig-nehrdich-2018-sanskrit}, vanilla \textbf{Transformer} \cite{vasvani_attention} and character-level Transformer with relative position encoding  \cite[\textbf{TENER}]{yan2019tener}. 
Finally, we consider lattice-structured approaches originally proposed for Chinese NER which incorporate lexical information in character-level sequence labelling architecture. 
These approaches consist of lattice-structured LSTM \cite[\textbf{Lattice-LSTM}]{zhang-yang-2018-chinese}, graph neural network (GNN) based architecture \cite[\textbf{Lattice-GNN}]{gui-etal-2019-lexicon} and Transformer based architecture \cite[\textbf{FLAT-Lattice}]{li-etal-2020-flat}.
\textbf{TransLIST:}  As per \S~\ref{LIST}, we report two variants: (a) TransLIST\textsubscript{ngrams} which makes use of only n-grams, and (b) TransLIST which makes use of SHR candidate space.
 \begin{table*}[!hbt]
\centering
\begin{adjustbox}{width=0.8\textwidth}
\small
\begin{tabular}{|c||c|c|c|c||c|c|c|c|}
\hline
\multirow{1}{*}{  } & \multicolumn{4}{c||}{\textbf{SIGHUM}} &\multicolumn{4}{c|}{\textbf{Hackathon}}\\\hline
\textbf{Model}                           & \textbf{P}     & \textbf{R}     & \textbf{F}    & \textbf{PM} & \textbf{P}     & \textbf{R}     & \textbf{F}    & \textbf{PM}   \\ \hline
Seq2seq                      & 73.44 & 73.04 & 73.24 & 29.20&72.31&72.15&72.23&20.21   \\ \hline
SupPCRW &76.30  & 79.47 & 77.85  & 38.64&-&-&-&-   \\ \hline
TENER                           & 90.03  & 89.20 & 89.61 & 61.24 & 89.38&87.33&88.35&49.92 \\ \hline
Lattice-LSTM                    & 94.36  & 93.83 & 94.09 & 76.99&91.47&89.19&90.31&65.76  \\ \hline
Lattice-GNN                     & 95.76 & 95.24 & 95.50 & 81.58&92.89&94.31&93.59&70.31  \\ \hline
Transformer                    & 96.52 & 96.21  & 96.36 & 83.88&95.79&95.23&95.51&77.70  \\ \hline
FLAT-Lattice                    & 96.75 & 96.70  & 96.72 & 85.65&\underline{96.44}&\underline{95.43}&\underline{95.93}&\underline{77.94}  \\ \hline
Cliq-EBM                      &96.18 & \underline{97.67} & \underline{96.92} & 78.83&-&-&-&-  \\ \hline 
rcNN-SS               & \underline{96.86} & 96.83  & 96.84 & \underline{87.08}&96.40&95.15&95.77&77.62   \\ \hline\hline
TransLIST\textsubscript{ngrams}  & 96.97 &96.77  &96.87  &86.52&96.68&95.74&96.21&79.28  \\ \hline
TransLIST  & \textbf{98.80} & \textbf{98.93} & \textbf{98.86} & \textbf{93.97}&\textbf{97.78}&\textbf{97.44}&\textbf{97.61}&\textbf{85.47}  \\ \hline
\end{tabular}
\end{adjustbox}
\caption{Performance evaluation between baselines in terms of P, R, F and PM metrics. The significance test between the best baselines, rcNN-ss, FLAT-lattice and TransLIST  in terms of recall/perfect-match metrics: $p < 0.05$ (as per t-test, for both the datasets).   We do not report the performance of SupPCRW and Cliq-EBM on Hackathon dataset due to unavailability of codebase.  On SIGHUM, we report numbers from their papers. The best baseline's results for the corresponding datasets are underlined. The overall best results per column are highlighted in bold.
}
\label{table:main_result}
\end{table*}
\paragraph{Results:}
\label{results}
Table~\ref{table:main_result} reports the results for the best performing configurations of all the baselines on the test set of benchmark datasets for the SWS task.\footnote{We do not compare with recently proposed variant of Clique-EBM \cite{amrith21} and seq2seq baseline \cite{aralikatte-etal-2018-sanskrit} due to unavailability of codebase. Also, they do not report performance on these two datasets.}  Except \textit{purely engineering} based systems (Seq2seq, TENER, Transformer and rcNN-SS), all systems leverage linguistically refined candidate solution space generated by SHR.
Among the lattice-structured systems, FLAT-Lattice demonstrates competing performance against rcNN-SS. 
We find that rcNN-SS and FLAT-Lattice perform the best among all the baselines on SIGHUM and Hackathon datasets, respectively.

Both the TransLIST variants outperforms all the baselines in terms of all the evaluation metrics with TransLIST providing an average 1.8 points (F) and 7.2 points (PM) absolute gain with respect to the best baseline systems, rcNN-SS (on SIGHUM) and FLAT-Lattice (on Hackathon). Even when the SHR candidate space is not available, the proposed system can use TransLIST\textsubscript{ngrams}, which provides an average 0.11 points (F) and 0.39 points (PM) absolute gain over the best baselines. TransLIST\textsubscript{ngrams} gives comparable performance to rcNN-SS on SIGHUM dataset, while on the Hackathon dataset, it performs significantly better than FLAT-Lattice ($p<0.05$ as per t-test). The wide performance gap between TransLIST and TransLIST\textsubscript{ngrams} demonstrates the effectiveness of using SHR candidate space, when available. 
Summarily, we establish a new state-of-the-art results with the help of meticulously stitched LIST, SMA and PRCP modules. The knowledge of the candidate space by SHR gives an extra advantage to TransLIST. Otherwise, natural choice is the proposed purely engineering variant TransLIST\textsubscript{ngrams} when that is not available.  
\begin{figure}[h]
\centering
  \subfigure[]{\includegraphics[width=0.235\textwidth]{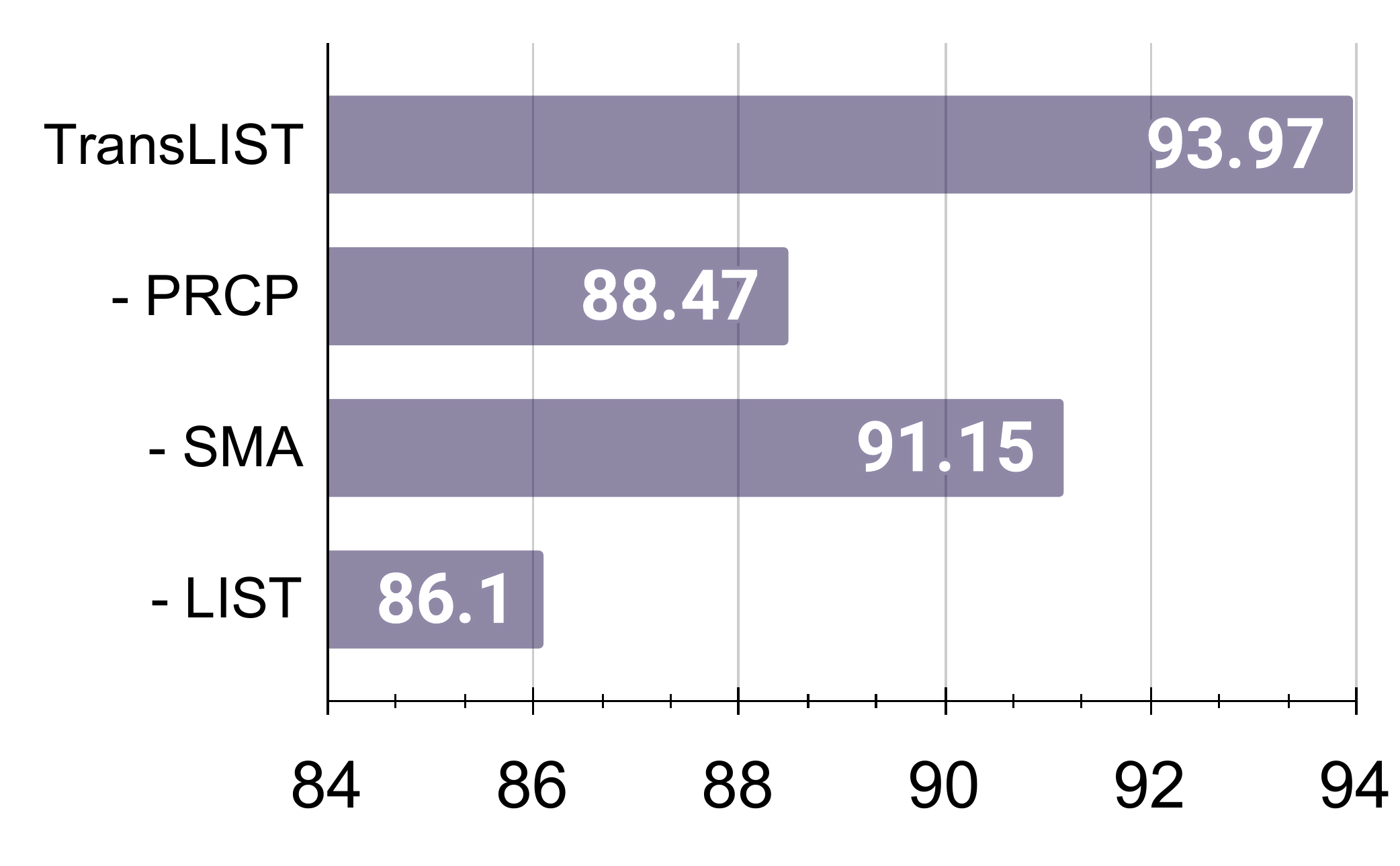}\label{fig:abl}} 
    \subfigure[]{\includegraphics[width=0.235\textwidth]{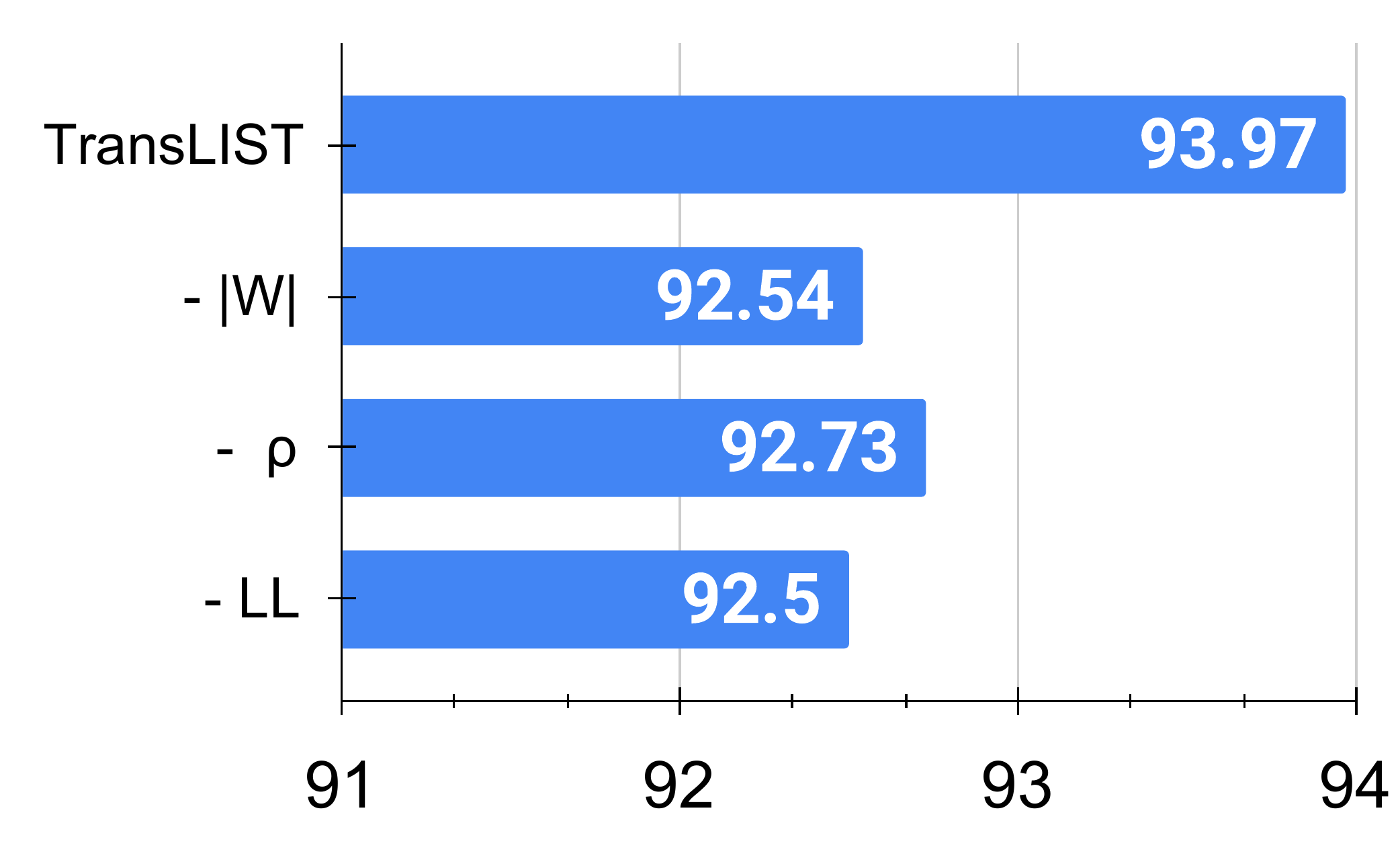}\label{fig:ci}} 
\caption{Ablations on (a) TransLIST (b) PRCP module in terms of PM (SIGHUM-test). Each ablation in (a) removes a single module from TransLIST. For example, ``-SMA'' removes SMA from TransLIST. For (b), ablations are shown by removing a particular term from path scoring function ($S$).} 
\label{fig:ablation} 
\end{figure}
\begin{figure*}[!htb]
\centering
\includegraphics[width=0.9\textwidth]{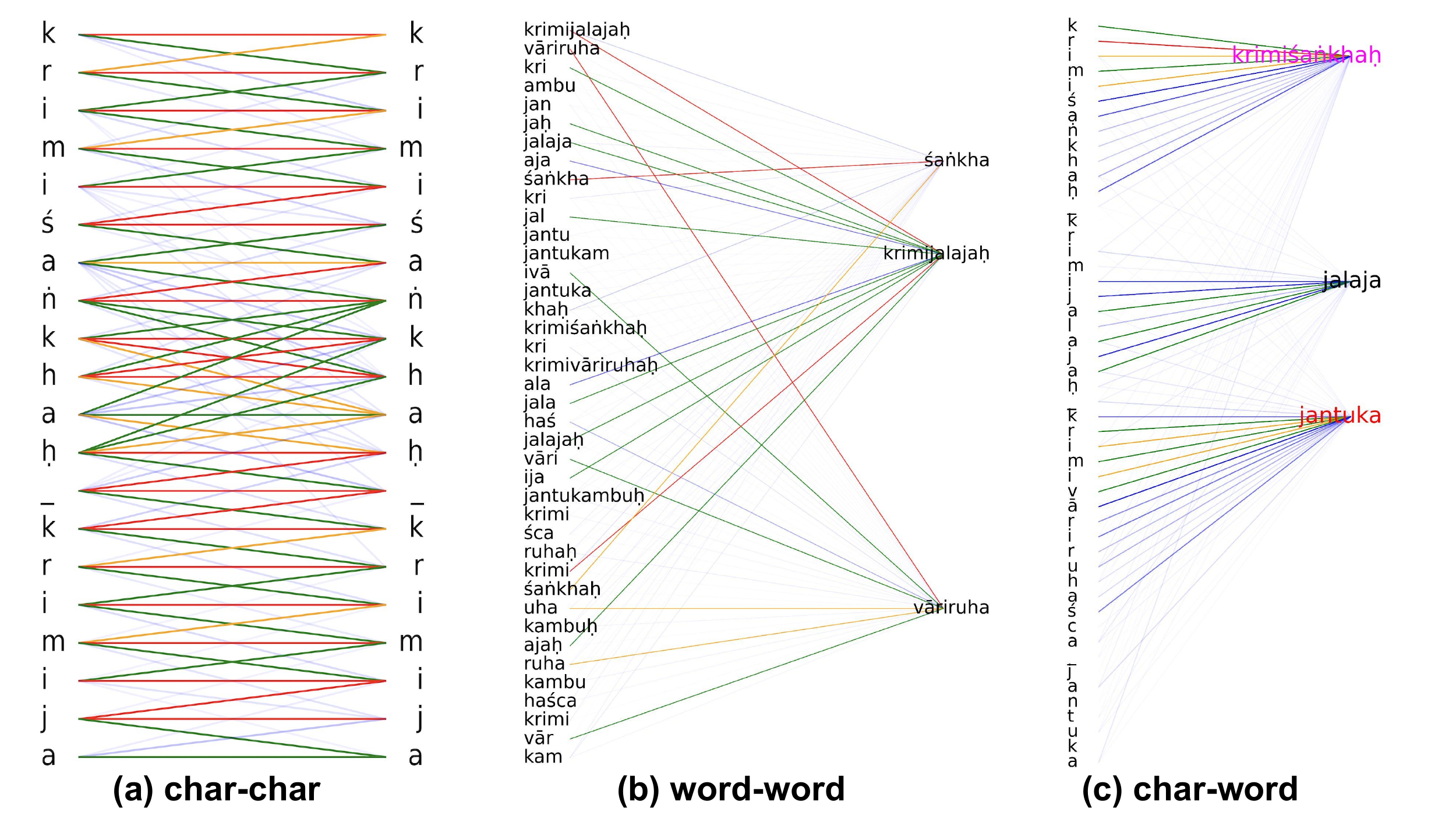}
\caption{SMA probing: Illustration of char-char, char-word and word-word interactions. The strength of the SMA decreases in the following order: red, orange, green and blue. Char-char attention mostly focuses on characters present in the vicinity of window size 1. Word-word interactions are able to capture whether a word is subword of another word or not. Finally, we find that quality of attention goes down for char-word as we move as per the following order: in vocabulary gold words (pink), in vocabulary non-golds (black) and out-of-vocabulary words (red). Some of the attentions are invisible due to very low attention score.} 
\label{fig:sma_probing} 
\end{figure*}

\section{Analysis}
\label{fine_grained_analysis}
In this section, we investigate various questions to dive deeper into the proposed components and investigate the capabilities of various modules. We use SIGHUM dataset for the analysis.
\noindent\paragraph{(1) Ablation analysis:} Here, we study the contribution of different modules towards the final performance of TransLIST. Figure~\ref{fig:abl} illustrates ablations in terms of PM when a specific module is removed from TransLIST. For instance, `-LIST' corresponds to character-level transformer encoder with SMA and PRCP. Removal of any of the modules degrades the performance. Figure~\ref{fig:abl} shows that LIST module is the most crucial for providing inductive bias of tokenization. Also, removal of `PRCP' module has a large impact on the performance. We observe that the PRCP module gets activated for 276 data points out of 4,200 data points in the test set.   We then deep dive into the PRCP path scoring function in Figure~\ref{fig:ci},  which consists of 3 terms, namely, penalty ($|W|$), perplexity score by CharLM ($|\rho|$) and log-likelihood ($LL$) by TransLIST, respectively. We remove a single term at a time from the path scoring function, and observe each of the terms used in the scoring function plays a major role in the final performance.

\noindent\paragraph{(2) Comparative analysis of potential LIST module variants to add inductive bias for tokenization:} 
We evaluate possible LIST variants which can help inject inductive bias for tokenization via auxiliary (word) nodes illustrated in Figure~\ref{fig:flat_architecture}: (a) \textit{sandhi} rules: We use \textit{sandhi} rules as a proxy to indicate potential modifications at specific position in the input sequence. For example, if input chunk contains the character \textit{`o'} (Figure~\ref{fig:sandhi_issue}) then it can be substituted with two possibilities \textit{\={o} $\rightarrow$ a-\={u}/a\d{h}}. We provide this proxy information through auxiliary nodes. (b) Sanskrit vocab: We obtain a list of vocabulary words from DCS corpus \cite{hellwig2010dcs} and add the words which can be mapped to the input character sequence using  a string matching algorithm. (c) n-grams: This is TransLIST\textsubscript{ngrams}  (d) SHR: We follow the exact settings as described in \S~\ref{LIST} except that we do not use the PRCP component. In Table~\ref{table:LIST_module}, we compare these with the \textit{purely engineering} variant of TransLIST (Base system: only character-level Transformer) where no inductive bias for tokenization is injected. Clearly, due to availability of enriched candidate space, SHR variant outperforms all its peers. However, competing performance of n-gram variant is appealing because it completely obliviates the dependency on SHR and remains unaffected in the absence of SHR's candidate space. 
\begin{table}[ht]
\centering
\begin{adjustbox}{width=0.45\textwidth}
\small
\begin{tabular}{|c|c|c|c|c|}
\hline
\textbf{System}                           & \textbf{P}     & \textbf{R}     & \textbf{F}  & \textbf{PM}    \\ \hline
Base system   & 92.75  & 92.62  &92.69 & 72.33 \\\hline
+\textit{sandhi} rules & 93.53 & 93.70 & 93.62 & 75.71  \\\hline
+Sanskrit Vocab & 96.75 & 96.70 & 96.72 & 85.65 \\\hline
+n-grams  &96.97 &96.77  &96.87  &86.52 \\\hline
+SHR & \textbf{97.79} & \textbf{97.45} & \textbf{97.62}& \textbf{88.47} \\ \hline
\end{tabular}
\end{adjustbox}
\caption{The comparison (on SIGHUM-test set) in between LIST variants. `+' indicates system where the corresponding variant is augmented with the base system. We do not activate PRCP for any of these systems.} 
\label{table:LIST_module}
\end{table}
\noindent\paragraph{(3) Probing analysis on SMA:} Here we analyze whether SMA upholds the prerequisite for effective modelling of inductive bias, i.e., prioritize candidate words which contain the input character for which the prediction is being made. Figure~\ref{fig:sma_probing} illustrates three types of interactions, namely, char-char, char-word and word-word. We use color coding scheme to indicate the strength of attention weight. The attention weight decreases in the following order: Red, Orange, Green and Blue. Char-char attention mostly focuses on characters present in the vicinity of window size 1. This local information is relevant to make decisions regarding possible {\sl sandhi} split. 
Word-word interactions are able to capture whether a word is subword of another word or not. Finally, for char-word attention, we find that quality of attention goes down as we move as per the following order: in vocabulary gold words (pink), in vocabulary non-golds (black) and out-of-vocabulary (unseen in training but recognized by SHR) gold words (red). While the drop in attention from in-vocabulary gold tokens to out-of-vocabulary gold tokens is expected, the drop in attention from gold tokens to non-gold tokens is desired. 
Thus, this probing analysis suggests that SMA module helps to improve intra/inter interactions between character/words and this substantiates the need of SMA module in TransLIST.
\noindent\paragraph{(4) How does TransLIST perform in a non-trivial situation where multiple \textit{sandhi} rules are applicable?}
In Table~\ref{table:sandhi_special}, we report the comparison  with rcNN-SS for a critical scenario of a \textit{sandhi} phenomenon. Table~\ref{table:sandhi_special} represents the possible \textit{sandhi} rules that generate the surface character \textit{\={a}}.  
Following \newcite{design_goyal}, the sandhi rewrite rules are formalized as $u|v \rightarrow f/x_{--}$    \cite{kaplan-kay-1994-regular} where $x, v, f \in \Sigma$ , and $u \in \Sigma^{+}$. $\Sigma$ is the collection of phonemes, $\Sigma^{*}$: a set of all possible strings over $\Sigma$, and $\Sigma^{+}$ = $\Sigma^{*} - \epsilon$.
For example, the potential outputs for the input ā can be \={a}, \={a}-\={a}, \={a}-a, a-a and a\d{h}. The correct rule can be decided based on the context.
These multiple rules pose a non-trivial challenge for a system to identify the applicability of specific rule. Therefore, it is interesting to compare the TransLIST with current state-of-the-art system to verify its ability for semantic generalization. We observe that TransLIST consistently outperforms rcNN-SS in terms of all metrics.\footnote{Follwing  \citet{hellwig-nehrdich-2018-sanskrit}, we report character-level F-score metric. $P = \frac{|S_{g} \cap S_{p}|}{|S_{p}|}$ ; $R = \frac{|S_{g} \cap S_{p}|}{|S_{g}|}$,  $F1 =\frac{2PR}{P+R}$,  ($S_{g}$) : Set of locations where the rule occurs in gold output, ($S_{p}$) : Set of locations where the rule is predicted.}  Table~\ref{table:sandhi_special} describes rules in decreasing order of their frequency. Interestingly, we notice large improvements over the current state-of-the-art system, especially for rare \textit{sandhi} rules. This observation confirms superior performance of TransLIST over the current state-of-the-art system.
\begin{table}[h]
\centering
\begin{adjustbox}{width=0.45\textwidth}
\small
\begin{tabular}{|c|c|c|c||c|c|c|}
\hline
\multirow{1}{*}{  } & \multicolumn{3}{c||}{\textbf{rcNN-SS}} &\multicolumn{3}{c|}{\textbf{TransLIST}}\\\hline
\textbf{Rules}                           & \textbf{P}     & \textbf{R}     & \textbf{F}    & \textbf{P}     & \textbf{R}     & \textbf{F}  \\ \hline
 
\={a}   & 99.3 & 99.3 & 99.3 & 99.7  & 99.6 & 99.6 \\\hline
a-a & 95.4 & 96.6 & 96.0 & 96.6  & 97.8 & 97.2 \\\hline
\={a}-a  & 88.4 & 83.1 & 86.5 & 90.5  & 83.8 & 87.0 \\\hline
\={a}\d{h} & 76.7 & 70.1 & 73.7 & 77.2  & 80.1 & 78.0 \\\hline
\={a}-\={a} & 50.1 & 42.1 & 45.7 & 80.0 & 40.9 & 53.3 \\ \hline
\end{tabular} 
\end{adjustbox}
\caption{The comparison (on SIGHUM-test set) in terms of P, R and F metrics between rcNN-SS and the TransLIST for ambiguous \textit{sandhi} rules leading to the same surface character \textit{\={a}}. The proposed model consistently outperforms rcNN-SS in all the metrics.} 
\label{table:sandhi_special}
\end{table}
 \begin{table*}[t]
\centering
\begin{adjustbox}{width=0.9\textwidth}
\small
\begin{tabular}{|c|c|c|}
\hline
&\textbf{Sentence}& \textbf{F-score}\\\hline
\textbf{Input sentence}& kimetadīśe bahuśobhamāne vā\d{m}bike yak\d{s}avapuścakāsti& -\\&\textit{Translation:} What is this body resembling a \textit{Yaksha} that glows, &\\
&oh Ambika! You who lord over! You who shine!& \\\hline 
\textbf{Correct segmentation}&kim etat īśe bahu śobhamāne vā ambike yak\d{s}a vapu\d{h} cakāsti &- \\ \hline
\textbf{SHR candidate space}& kim, etat, īśe, bahu, śobhamāne, śobham, āne, śobha, māne, &- \\
&mā, vā, ambike, yak\d{s}a, vapu\d{h}, cakāsti, ca, kā, asti&\\
&\textit{Word-word meaning:} what, this, the one who lord, very much,&\\
& the one who shine, bright, mouth, I respect, never, or, Parvati, &\\
&a kind of celestial being, body, glows, and, who (female), is there (be).&\\  \hline \hline
\textbf{rcNN-SS} &kim etat īśe bahu śobhamāne \textbf{vāmbike yak\d{s}avapu\d{h} caka asti}  & 52.60 \\\hline 
\textbf{TransLIST-PRCP} &kim etat īśe bahu śobhamāne vā \textbf{aambike} yak\d{s}a vapu\d{h} cakāsti  & 90.00\\\hline 
\textbf{TransLIST} &kim etat īśe bahu śobhamāne vā ambike yak\d{s}a vapu\d{h} cakāsti  & 100.00 \\\hline
\end{tabular}
\end{adjustbox}
\caption{An example to illustrate the effectiveness of PRCP module of TransLIST. Bold represents incorrect segmentation for the input sequence.} 
\label{table:case_study}
\end{table*}
 \begin{figure}[h]
    \centering
    \includegraphics[width=0.4\textwidth]{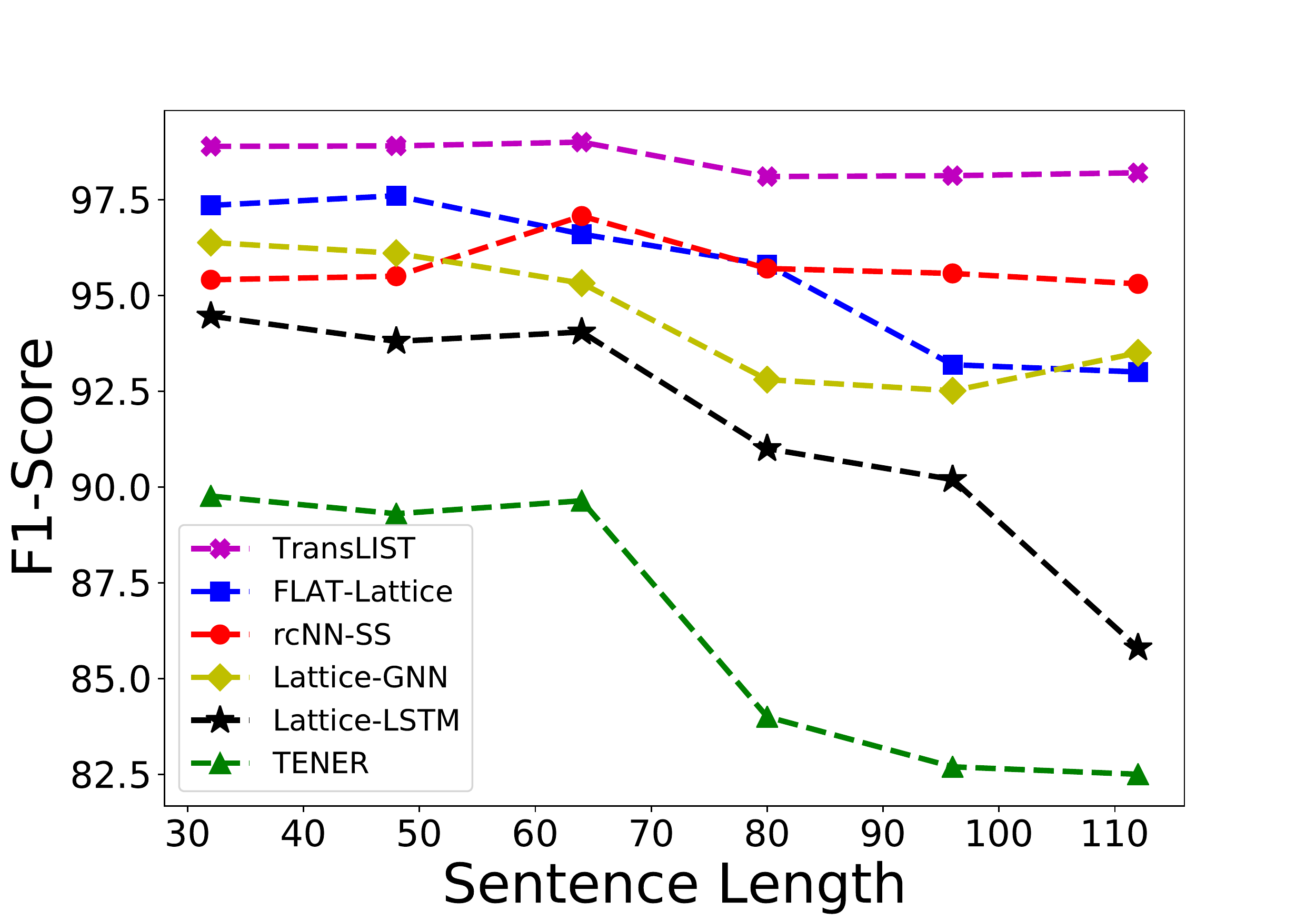}
    \caption{F1-score against sentence length (no. of characters) over the SIGHUM dataset}
    \label{fig:error_analysis}
\end{figure}
\noindent\paragraph{(5) How robust is the system when sentence length is varied?}
 In Figure~\ref{fig:error_analysis}, we analyze the performance of the baselines with different sentence lengths. 
We plot the F1-score against sentence length. Clearly, while all the systems show superior performance for shorter sentences, TransLIST is much more robust for longer sentences compared to other baselines. The lattice-structured baselines give competing F1-scores over short sentences but relatively sub-par performance over long sentences.

\noindent\paragraph{(6) Illustration of PRCP with an example:} 
Table~\ref{table:case_study} illustrates  an example that probes the effectiveness of PRCP in TransLIST. We compare TransLIST with rcNN-SS and observe that TransLIST also predicts words out of candidate solution space when PRCP module is not activated. However, the degree of such mistakes in TransLIST is comparatively less due to effective modelling of inductive bias for tokenization using LIST and SMA modules. In Table~\ref{table:case_study}, rcNN-SS predicts three words which are not part of candidate space, namely, \textit{vāmbike, yak\d{s}avapu\d{h}} and \textit{caka}. These are mistakes that can be rectified with the help of available candidate space.
Interestingly, TransLIST commits only a single mistake in this category by predicting out of solution space word \textit{aambike}. PRCP aids in mitigating such mistake by appropriately substituting suitable candidates. 
\section{Related Work}
\label{related_section}
Earlier approaches on SWS focused on rule-based Finite State Transducer systems \cite{gerard2003lexicon,mittal-2010-automatic}.
\citet{natarajan} attempted to solve the SWS task for sentences with one or two splits using the Bayesian approach. Recently, \citet[SHR]{goyal2016} proposed a \textit{lexicon driven} shallow parser.
This, along with the recent upsurge in segmentation datasets \cite{krishna-etal-2017-dataset,hellwig-nehrdich-2018-sanskrit,hackathon_data} led to two categories of approaches, namely, \textit{lexicon driven} \cite{krishna-etal-2016-word,krishna-etal-2018-free,amrith21} and \textit{purely engineering} \cite{hellwig2015using,hellwig-nehrdich-2018-sanskrit,aralikatte-etal-2018-sanskrit,reddy-etal-2018-building}.
These existing approaches for SWS are either brittle in realistic scenarios or do not consider the potentially useful/available information.
Thus, TransLIST bridges the shortcomings exhibited by each family and gives a win-win solution that marks a new state-of-the-art results.

\section{Conclusion and Discussion}
\label{conclusion}
 In this work, we focused on Sanskrit word segmentation task. To address the shortcomings of existing \textit{purely engineering} and \textit{lexicon driven} approaches, we demonstrate the efficacy of TransLIST as a win-win solution over drawbacks of the individual lines of approaches. TransLIST induces inductive bias for tokenization in a character input sequence using the LIST module, and prioritizes the relevant candidate words with the help of soft-masked attention (SMA module). Further, we propose a novel path ranking algorithm to rectify corrupted predictions using linguistic resources on availability (PRCP module).
Our experiments showed that TransLIST provides a significant boost with an average 7.2 points (PM) absolute gain compared to the best baselines, rcNN-SS (SIGHUM) and FLAT-Lattice (Hackathon). We have also showcased fine-grained analysis on TransLIST's inner working. 
We plan to extend this work for morphological tagging in standalone mode \cite{gupta-etal-2020-evaluating} and multi-task setting  \cite{krishna-etal-2018-free} with the SWS task. 

\section*{Limitations}
The preliminary requirement to extend TransLIST for the languages which also exhibit {\sl sandhi} phenomenon is {\sl lexicon-driven} shallow parser similar to Sanskrit Heritage Reader (SHR). Otherwise, the natural choice is the proposed purely engineering variant TransLIST\textsubscript{ngram}. It would be interesting to check if TransLIST and TransLIST\textsubscript{ngram} can be used together.

\section*{Ethics Statement} We do not foresee any ethical concerns with the work presented in this
manuscript.

\section*{Acknowledgements}
We are grateful to Oliver Hellwig for providing the
DCS Corpus and Gerard Huet for providing the Sanskrit Heritage Engine. We thank Sriram Krishnan, University of Hyderabad and Hackathon organizers\footnote{\url{https://sanskritpanini.github.io/index.html}} for providing Hackathon dataset. We thank Amrith Krishna, University of Cambridge for clarifying our queries related to SIGHUM dataset and evaluation metrics.  We are grateful to Rishabh Kumar, IIT Bombay for helping us with evaluation of Cliq-EBM baseline.
We would like to thank the anonymous reviewers for their constructive feedback towards improving this work. The work of the first author is supported by the TCS Fellowship under the Project
TCS/EE/2011191P.

\bibliography{anthology,custom}
\bibliographystyle{acl_natbib}

\appendix

\section{Appendix}
 \paragraph{Average run times:} Table~\ref{table:run_time} shows the average training time in hours and inference time in milliseconds for all competing baselines. We find that pure engineering-based techniques (TENER, rcNN-SS) outperform lattice-structured architectures (Lattice-LSTM, Lattice-GNN, FLAT-Lattice) in terms of run time. When the inference times of TransLIST and TransLIST\textsubscript{ngrams} are compared, TransLIST takes longer owing to the PRCP module. It would be interesting to explore approaches to optimise the inference time of the PRCP module.
 
  \begin{table}[h]
\centering
\resizebox{0.4\textwidth}{!}{
\begin{tabular}{|c|c|c|}
\hline
\textbf{System}                            & \textbf{Train (Hours)}  & \textbf{Test (ms)}  \\ \hline

TENER                           & 4 H & 7 ms\\ \hline
Lattice-LSTM                    & 16 H  & 110 ms\\ \hline
Lattice-GNN                     & 64 H & 95 ms \\ \hline
FLAT-Lattice                    & 5 H & 14 ms\\ \hline
rcNN-SS               & 4 H & 5 ms\\ \hline
Cliq-EBM                      &10.5 H & 750 ms \\ 
 \hline 
 TransLIST\textsubscript{ngrams}  & 8 H & 14ms \\ \hline
TransLIST  & 8 H & 105 ms\\ \hline
\end{tabular}}
\caption{Average training time (in hours) and inference time (in milliseconds)  for all the competing baselines.}
\label{table:run_time}
\end{table}
\end{document}